# Physics-driven Synthetic Data Learning for Biomedical Magnetic Resonance

Qinqin Yang[1], Zi Wang[1], Kunyuan Guo[1], Congbo Cai[1], and Xiaobo Qu[1,*]

## Abstract

Deep learning has innovated the field of computational imaging. One of its bottlenecks is unavailable or insufficient training data. This article reviews an emerging paradigm, imaging physics-based data synthesis (IPADS), that can provide huge training data in biomedical magnetic resonance without or with few real data. Following the physical law of magnetic resonance, IPADS generates signals from differential equations or analytical solution models, making the learning more scalable, explainable, and better protecting privacy. Key components of IPADS learning, including signal generation models, basic deep learning network structures, enhanced data generation and learning methods are discussed. Great potentials of IPADS have been demonstrated by representative applications in fast imaging, ultrafast signal reconstruction and accurate parameter quantification. Finally, open questions and future work have been discussed.

## Index Terms

Data synthesis, Physical model, Biomedical magnetic resonance, Deep learning

## I. INTRODUCTION

Data learning has empowered the computational imaging with fast sampling, ultrafast signal reconstruction and straightforward parameter quantification [1-4]. A milestone work of deep learning fast MRI was presented in [3]. In the age of deep learning, a large amount of high-quality data is essentially important to achieve excellent performance. However, in biomedical imaging, these data may be hard to be acquired in the challenging applications, e.g., the blurred images of moving organs, the lengthy measuring of quantitative physical parameters, or the irreversible acquisition of physiological processes. Thus, new data generation and learning schemes are highly desired to boost biomedical imaging applications.

[1] Department of Electronic Science, Biomedical Intelligent Cloud Research and Development Center, Fujian Provincial Key Laboratory of Plasma and Magnetic Resonance, National Institute for Data Science in Health and Medicine, Xiamen University, Xiamen, China, 361005
\* Corresponding author (quxiaobo@xmu.edu.cn)

Recently, synthetic data starts to attract attention in computational imaging [5, 6]. Learning in synthetic data could reduce the dependence on paired real-world data, quickly generate massive data, overcome the difficulties or even the impossibility to collect real data, and protect privacy in biomedicine. Here, we focus on the imaging physics-based data synthesis (IPADS) because it follows plausible physical model and enables good interpretability [5]. To make the discussions compact, we limit the content to the biomedical Magnetic Resonance (MR) since IPADS learning (Fig. 1) has become frontiers in this area, such as fast quantitative imaging [7-19], signal reconstruction [20-26] and pulse sequence optimization [27, 28]. The concept of IPADS learning could be generalized to other computational imaging modalities as long as an appropriate physical model and learning network are included.

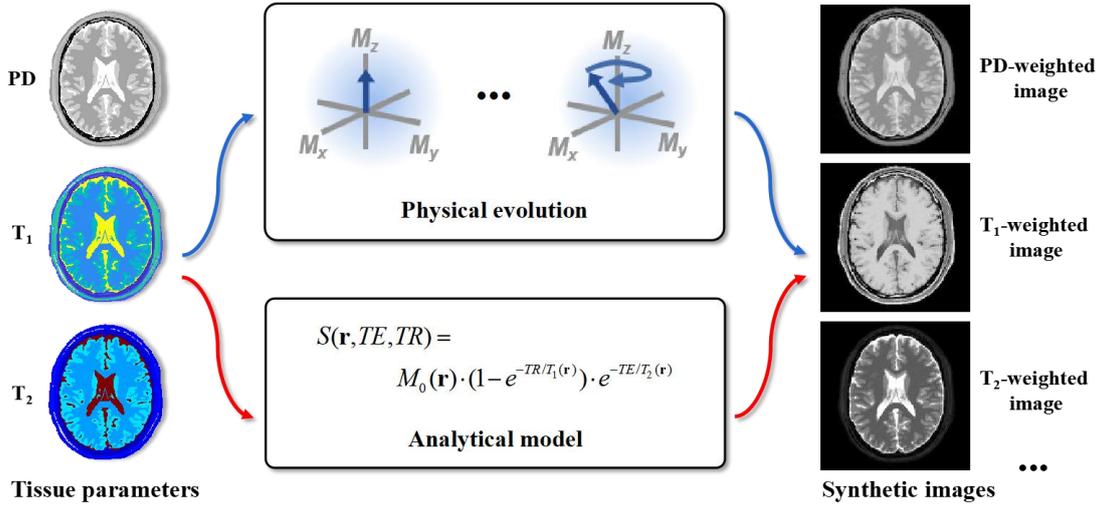

**Fig. 1**. Two lines of imaging physics-based data synthesis (IPADS) in biomedical magnetic resonance. Various synthetic images can be generated from numerical tissue parameters through physical evolution or analytical models. Note: PD is short for proton density.

This article will give an overview of IPADS for deep learning magnetic resonance. Based on the physical signal evolution or not, we first divide the IPADS into two lines, including the physical signal evolution and analytical modeling (Fig. 1), and then discuss the enhanced learning with realistic data adaption and advanced network structures. Representative applications and future works will be provided.

## II. MAGNETIC RESONANCE PHYSICS

The physics of MR governs signal formation, which involves spin dynamics, quantum mechanics and electromagnetism. For example, MR imaging (MRI) presents spatially structural information with different contrasts, while MR spectroscopy (MRS) focuses on the spectral signals of multiple molecules. The MR signal evolution may have analytical solutions under some conditions or may not have them in general. Accordingly, IPADS methods are divided into three categories and some representative works are summarized in Table 1.

Table 1. A summary of IPADS in biomedical magnetic resonance

| Category | Data generation | Learning issues | Method |
|---|---|---|---|
| Physical evolution | Bloch simulation | MR fingerprinting | DRONE [7] |
| | Bloch simulation | $T_2$ mapping | OLED [9] |
| | Bloch simulation | Water-fat separation | SPEN-WFS [25] |
| | Bloch simulation | Sequence optimization | MRzero [27] |
| | Bloch simulation | Cardiac motion tag tracking | SyntheticCMR [26] |
| | Bloch simulation + FDTD | Electrical properties tomography | DL-EPT [13] |
| Analytical model | Dipole convolution model | Quantitative susceptibility mapping | QSMnet+ [15] |
| | Multi-pool exponential model | Myelin water fraction estimation | ANN-MWF [17] |
| | Bloch-McConnell equation | CEST z-spectra quantification | ANNCEST [18] |
| | Lorentzian line shape model | Spectra reconstruction | DL-NMR [20] |
| Hybrid approach | Quantum mechanical simulation + Exponential signal model | MRSI reconstruction | DL-MRSI [23] |
| | Bloch simulation + Dipole convolution model | SPIO mapping | DeepSPIO [16] |

FDTD: finite difference time domain method; DL: deep learning; MRSI: MR spectroscopic imaging; CEST: chemical exchange saturation transfer; RF: radio frequency; QSM: quantitative susceptibility mapping; EPT: electrical properties tomography; SPIO: super paramagnetic iron oxide.

*A. Physical Evolution*

Bloch equation [29] describes the time-dependent evolution of magnetization vector $\mathbf{M}(t) \in \mathbf{R}^3$ under an external magnetic field $\mathbf{B}(t) \in \mathbf{R}^3$. Two key phenomena in magnetized spin, i.e., precession and relaxation, can be formulated in the differential equations:

$$\frac{d\mathbf{M}(t)}{dt} = \underbrace{\gamma \mathbf{B}(t) \times \mathbf{M}(t)}_{\text{Precession}} + \underbrace{R(\mathbf{M}(t))}_{\text{Relaxation}}, \tag{1}$$

where γ is the gyromagnetic ratio (e.g., 42.6 MHz/T for hydrogen nuclei). For the precession term, the motion of spins is determined by many factors, including the main static field $B_0$, the radio-frequency (RF) pulse field $B_1(t)$, the chemical shift (frequency) $\Delta\omega_0$ and the linear gradient fields $\mathbf{G}(t)$, at location $\mathbf{r}$ as:

$$\mathbf{B}(t) = (B_0 - \frac{\Delta\omega_0}{\gamma} + \mathbf{G}(t) \cdot \mathbf{r})\hat{z} + \mathbf{B}_1(t), \tag{2}$$

where the thermal equilibrium magnetization vector is tipped from the $B_0$ direction (z) into the transverse plane (x-y) under the effect of $B_1(t)$. The $\mathbf{G}(t)$ is usually used to modulate phase or encode different spatial locations. In contrast, the relaxation term describes the process of magnetization vector returning to its equilibrium state. Two relaxation time constants, $T_1$ and $T_2$, are used to characterize the regrowth of longitudinal magnetization ($M_z$) and the decay of the transverse magnetization ($M_{x,y}$), respectively. These relaxation parameters are very valuable in clinics, such as lesion diagnosis [12].

MR pulse sequence describes a series of physical radio frequency pulses applied to the objects, resulting in a particular image or spectrum appearance. It usually consists of a series of varying $B_1(t)$ and $G(t)$ in the form of timing diagram. Typically, the signal model for a simple MR pulse sequence has an analytical solution if steady-state is assumed. However, as the complexity of the pulse sequence increases, analytical solutions are hard to obtain due to spin history effects at unsteady-states and system imperfections. For example, in MR fingerprinting [7, 30], various sequence components are varied in a pseudorandom pattern and MR signals are not analyzed using the analytic expression but the dictionary matching.

*B. Analytical Model*

Analytical model provides a clear closed form solution of MR signals and could approximate the physical evolution of MR signal under some assumptions or simplifications.

In MRI, consider a most common pulse sequence, the spin-echo sequence, Eq. (3) provides an analytical solution for image contrast as:

$$S(\mathbf{r}, TE, TR) = M_0(\mathbf{r}) \cdot (1 - e^{-TR/T_1(\mathbf{r})}) \cdot e^{-TE/T_2(\mathbf{r})}, \qquad (3)$$

assuming that the initial magnetization vector undergoes the action of 90° and 180° radio frequency pulses, and the repetition time is much larger than echo time. $M_0(\mathbf{r})$ represents equilibrium longitudinal magnetization. By adjusting the repetition and echo time, image contrasts can be generated if the tissue parameters $M_0(\mathbf{r})$, $T_1(\mathbf{r})$ and $T_2(\mathbf{r})$ are provided.

In addition to relaxation, magnetic susceptibility $\chi(\mathbf{r})$ represents the ability of a substance to become magnetized under an applied magnetic field. Tissue-specific magnetic susceptibility variations within the MRI scanner can cause inhomogeneity of the static magnetic field $B_0$. To describe the field variations caused by $\chi(\mathbf{r})$, a dipole convolution

model is typically used as [14-16]:

$$\Delta B_0(\mathbf{r}) = B_0 \cdot \chi(\mathbf{r}) * D(\mathbf{r}), \qquad (4)$$

in which, the induced field inhomogeneity $\Delta B_0(\mathbf{r})$ is expressed as the convolution between the spatial distribution of susceptibility $\chi(\mathbf{r})$ and a unit dipole response $D(\mathbf{r})$.

Not limited to the image, in MRS, the spectral signal of each individual voxel comes from multiple molecules. This signal is commonly modeled as [20-22, 31]:

$$S(\mathbf{r},t) = \sum_{m=1}^{M} c_m(\mathbf{r}) v_m(t) e^{-t/T_{2,m}(\mathbf{r})}, \qquad (5)$$

where the $m$ denotes the $m^{\text{th}}$ molecule, $c_m(\mathbf{r})$ and $v_m(t)$ are its concentration and basis function, respectively. For the *ex vivo* biological MRS, which are used to determine the concentrations of metabolites or structures of proteins, the objects (usually in liquid or solid) to be acquired are placed in a tube and treated as a whole. Thus, the spatial location *r* is commonly ignored and only the $S(t)$ is acquired from scanners. In biological MRS, the basis function could be expressed as a linear combination of multiple exponentials ($J_m$) as [20-22, 31]:

$$v_m(t) = \sum_{j=1}^{J_m} (a_{j,m} e^{i\phi_{j,m}}) e^{i2\pi f_{j,m} t}, \qquad (6)$$

where $i$ is the imaginary unit, $a_{j,m}, f_{j,m}$ and $\phi_{j,m}$ are the amplitude, frequency, and phase of the $j^{th}$ spectral peak, respectively. By performing the Fourier transform on $S(t)$, a spectrum will be obtained and the spectral peaks follow the Lorentzian line shape [20-22, 31].

For *in vivo* MRS, Eq. (5) is sub-optimal since it does not consider imperfect but real imaging conditions, e.g., field inhomogeneity and motions. More practically, the signal of MRS could be modeled as [23, 24]:

$$S(\mathbf{r},t) = \sum^{M} c_m(\mathbf{r}) v_m(t) e_m(t; \theta_m(\mathbf{r})) + b(\mathbf{r},t), \qquad (7)$$

where $b(\mathbf{r},t)$ is a baseline signal mainly contributed by macromolecules and commonly follows the Gaussian line shape, $e_m(t; \theta_m(\mathbf{r}))$ captures molecule-dependent time-domain modulation functions that can be described by some experimental and physiological parameters in $\theta_m(\mathbf{r})$.

## III. SYNTHETIC DATA LEARNING

### A. Physical Evolution

Physical evolution-based IPADS (PE-IPADS) and learning relies on simulating discrete spin motion at small time interval using the Bloch equation. This process can be described with successive operators according to a specific MR pulse sequence. The signal formation, however, is computationally expensive, which involves large-scale matrix operations, integration and differentiation. Fortunately, several MR physical simulation tools have been developed, e.g., SPROM[1] [32], JEMRIS[2], MRiLab[3] [33]. With these tools, one has to set proper physical parameters to make data generation in IPADS as realistic as possible.

Physical parameters include object-specific and experiment-specific parameters. The former indicates the nature of the scanned objects, such as the proton density, relaxations ($T_1$, $T_2$ and $T_2^*$), diffusion and electromagnetic properties at a particular spatial location in MRI, or amplitudes, resonance frequencies and metabolite concentrations in MRS. The latter takes the pulse sequence and real imaging conditions into account, such as the repetition time, echo time, flip angle, the strength of the main field magnetic field ($B_0$) and its inhomogeneity ($\Delta B_0$), radio frequency field inhomogeneity ($B_1$), and eddy currents, etc.

Up to now, most PE-IPADS learning is on imaging [7-13, 25-27], especially quantitative parametric imaging [7-13]. A representative application of PE-IPADS learning is the $T_2$ mapping with overlapping-echo detachment (OLED) [10]. OLED is an ultrafast imaging sequence that encodes information of multiple image into one image, allowing parametric imaging within very short time (within 10 s for a whole brain). How to estimate reliable quantitative parameters within the same imaging time is a main problem. Deep learning has been evidenced powerful but the lack of real-world pairs of quantitative parameters and images should be addressed.

A typical flow of data generation for OLED is illustrated in Fig. 2. First, object-shape templates should be provided. They are common created by randomly filling blank templates with hundreds of different 2D/3D basic geometric shapes [9-11, 14, 16, 19, 25, 26]. Second, tissue-specific parameters, such as the proton density and $T_2$ relaxation at particular spatial locations are set in the ranges of [0, 1] and [20, 700] ms, respectively. Then, pulse sequence needs to be programmed and imaging parameters (echo times= 22, 52. 82, 110 ms, flip angle=30°) are set to be consistent with real imaging experiments. Last, an imperfection imaging condition, the radio frequency field inhomogeneity ($B_1$), is generated by random polynomial functions.

---

[1] https://doi.org/10.6084/m9.figshare.19754836.v2
[2] https://www.jemris.org
[3] http://mrilab.sourceforge.net

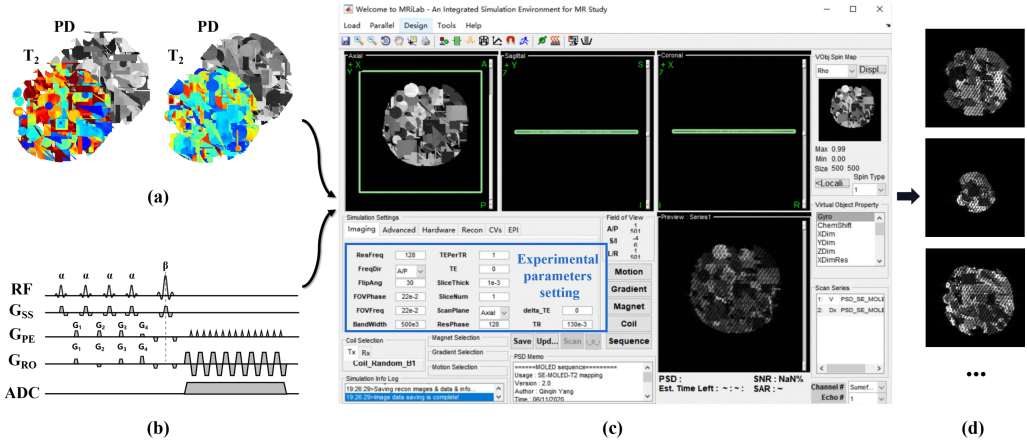

**Fig. 2**. An illustration of physical evolution-based data generation. (a) Parametric templates that have geometric shapes and object-parameters, (b) Magnetic resonance pulse sequence needs to be programmed according to the real-world experiment, (c) A software (here is the MRiLab [33]) that enables physical signal evolution, (d) Massive synthetic data.

Once the data is generated with physical simulation tools, a network structure should be designed to conduct the learning with PE-IPADS data. A direct and common chosen approach is to learn the mapping from the signal to quantitative parameters in the end-to-end way [7, 9, 13]. For example, the OLED images were mapped to $T_2$ parameters in Fig. 3. In general, for a deep learning network $\mathcal{N}$ with trainable network parameter $\boldsymbol{\theta}$, an optimal learning is to minimize the loss $L(\cdot)$ of estimated quantitative object-parameters $\hat{\mathbf{p}}_n = \mathcal{N}(\mathbf{s}_n; \boldsymbol{\theta})$ according to:

$$\hat{\boldsymbol{\theta}} = \arg\min_{\boldsymbol{\theta}} \sum_{n=1}^{N} L\left(\mathbf{p}_n - \mathcal{N}(\mathbf{s}_n; \boldsymbol{\theta})\right), \qquad (8)$$

where $\mathbf{p}_n$ is the ground-truth but simulated quantitative parameters, $\mathbf{s}_n$ is the generated images, and $n$ denotes the $n^{\text{th}}$ sample and its total number is $N$. Thus, PE-IPADS learning tries to approximate the inverse process from physical quantitative parameters of scanned objects to the output signal.

Once the network is trained with sufficient samples, estimating quantitative parameters $\mathbf{p} = \mathcal{N}(\mathbf{s}; \hat{\boldsymbol{\theta}})$ from a target image $\mathbf{s}$ becomes a forward and fast process. For the PE-IPADS learning for OLED imaging, 800 images were trained in around 22 hours and less than 1 second was consumed to obtain faithful $T_2$ maps (Figs. 3(b) and (c)). High fidelity was achieved on the correlation coefficients (0.999 and 0.997 in phantom and human brain data) to the $T_2$ maps of conventional imaging pulse sequence, but the data acquisition time is reduced from 17 minutes to 10 seconds [10]. Besides, OLED with PE-IPADS avoids challenging motion artifacts in an epilepsy patient (Fig. 3(d)) [12].

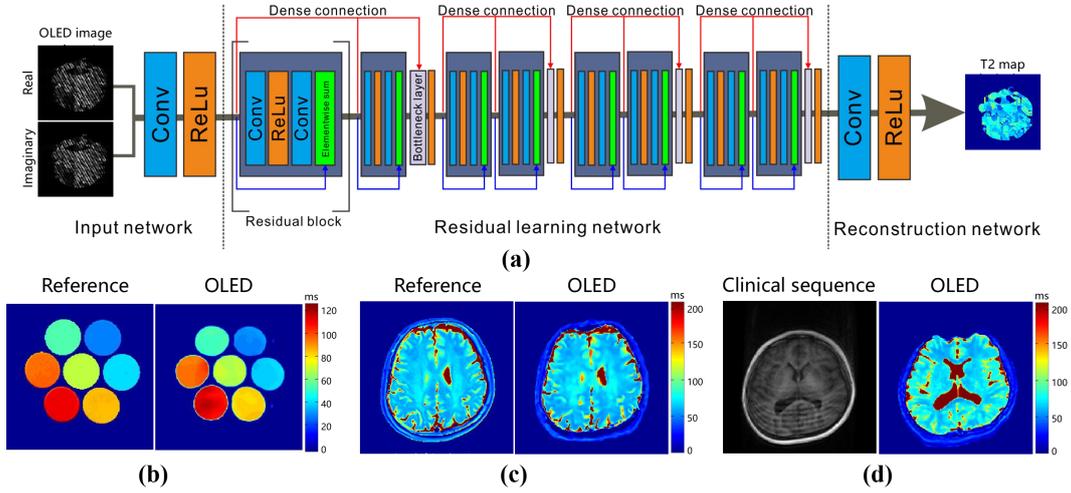

**Fig. 3**. Residual dense network structure for $T_2$ mapping with OLED imaging sequence [10, 12]. (a) The network architecture learns the end-to-end mapping from the images to $T_2$ parameters, (b)-(c) Reconstructed $T_2$ maps of OLED and the corresponding reference of conventional acquisition in phantom and *in vivo* brain data, respectively. (d) An image of clinical sequence and $T_2$ map of OLED from an epilepsy patient.

The PE-IPADS has been applied to many other imaging scenarios due to its high flexibility [25-28]. At the current early stage research, the network structure seems not the primary focus. Most approaches adopted mainstream network structures, e.g., fully connected network in MR fingerprinting [7, 8] and residual network in cardiac motion tag tracking [26]. Through simulating the imaging process of water and fat, the SPEN-WFS [25] trained a U-Net on the generated data and fast separated water and fat in 0.46 s on a personal computer (tradition method costs 30.31 s). Additionally, with a 3D convolution neural network, the common Bloch equation evolution could be combined with electromagnetic simulation to obtain coil-specific radio frequency profiles and phases [13]. This approach leverages PE-IPADS into electrical property tomography, extending traditional MRI to new imaging modalities.

*B. Analytical Model*

Analytical model-based IPADS (AM-IPADS) skips the complex physical evolution process. It directly and quickly generates a large amount of training data by adjusting the physical parameters in the closed form solution expression [14, 15, 17-22]. It is interesting to see that most AM-IPADS learning is on spectra in MRS [18-22] and few applications on MRI [14, 15, 17]. Thus, in the following, we mainly discuss the spectra applications.

In *ex vivo* biological MRS, the spectral peaks are usually in Lorentzian lineshapes and the effect of system imperfection is negligible in the ideal scenario, i.e., only object-specific parameters need to be considered. Specifically, synthetic data are generated using general exponential functions [20-22].

The AM-IPADS are commonly used in the spectra reconstruction [20-22] that does not involve the physical evolution process. Reconstruction aims at estimating a high-quality spectrum from the undersampled or low signal-to-noise data. Although deep learning has shown astonishing performance in image reconstruction of MRI, the methodology was developed relatively later in MRS due to the lack of paired realistic data. IPADS relaxes this requirement through data generation according to Eqs. (5)-(7). As the physical parameters, such as amplitudes, resonance frequencies and metabolite concentrations can be simulated, IPADS strongly increases the flexibility of the deep learning. In general, the network $\mathcal{N}$ learns the trainable parameters $\boldsymbol{\theta}$ to minimize the total difference between the fully-sampled (or noise-free) label signal $\tilde{\mathbf{s}}$ and the output of network $\mathcal{N}(\mathbf{d}_n;\boldsymbol{\theta})$ as follows:

$$\hat{\boldsymbol{\theta}} = \arg\min_{\boldsymbol{\theta}} \sum_{n=1}^{N} L\left(\tilde{\mathbf{s}}_n - \mathcal{N}(\mathbf{d}_n;\boldsymbol{\theta})\right), \qquad (9)$$

where $N$ is the number of training samples, $L$ is the loss function such as the $l_2$ norm loss. After obtaining the optimal network parameters $\hat{\boldsymbol{\theta}}$, a target signal $\mathbf{s}$ is reconstructed via $\mathbf{s} = \mathcal{N}(\mathbf{d};\hat{\boldsymbol{\theta}})$ for a given undersampled (or noisy) realistic input $\mathbf{d}$. The reconstruction is always ultrafast in realistic experiments, e.g., only 0.04/2.75 seconds for reconstructing

2D/3D protein spectra and is about 30 times faster than the conventional compressed sensing methods [22]. The basic network architectures are convolutional neural networks (CNN) [22] or its densely connected version [20, 21].

Without any real data involved in training, a representative application of the AM-IPADS learning is ultrafast MRS reconstruction [20-22]. Firstly, based on the exponential model in Eq. (5), we vary spectral parameters according to the uniform distribution, such as discretely randomizing the number of peaks from 1 to 10, normalized amplitude from 0.05 to 1, and the normalized frequency from 0.01 to 0.99 Hz [20-22]. In total, 40000 pairs of synthetic data (inputs are undersampled time-domain free induction decay, FID, signals and output labels are fully sampled spectra) are generated within several seconds. Then, a deep learning magnetic resonance network, DLNMR [20], is trained with the synthetic data in 5~31 hours. The flowchart of DLNMR (Fig. 4(a)) shows that the spectrum artifacts introduced by undersampling are first removed with dense CNN and then the intermediate spectra are further refined to maintain the data consistency to the sampled FID. With the increase of the network phase, artifacts are gradually removed, and finally a clean spectrum can be reconstructed. The DLNMR is good at restoring high-intensity peaks (the 2$^{nd}$ row of Fig. 4(c)). It has been applied to reconstruct many real spectra of proteins, achieving the peak correlation up to 0.9996 in 2D spectra and the acceleration factor of sampling up to 10 in 3D spectra. Even though, small peaks (the 2$^{nd}$ row of Fig. 4(c)) may be comprised and even becomes worse (the 2$^{nd}$ row of Fig. 4(d)) if the mismatch is existed between the training and target data. These observations imply that, without any real data in training, AM-IPADS has great protentional to boost deep learning but still needs further improvements.

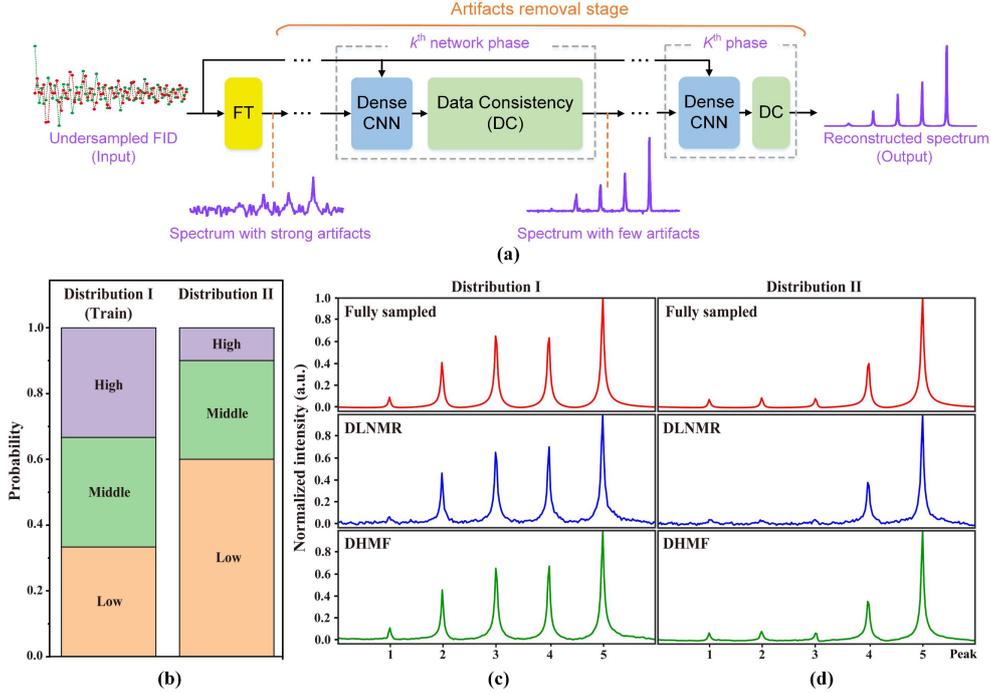

**Fig. 4**. Exponential signal reconstructions with mismatched training data [20, 21]. (a) Recursive DLNMR framework that alternates between the dense CNN and the data consistency. (b) A uniformly distribution I of peak intensities for training or reconstruction and a non-uniform distribution II that is only for reconstruction, (c) and (d) are reconstructed target signals that satisfy uniform and non-uniform distributions, respectively. Note: The sampling rate is 25%. The network structure of DHMF imitates the iterative process of low-rank matrix factorization MRS reconstruction.

In addition to signal reconstruction, AM-IPADS deep learning has been explored in physical parameter quantification, such as quantitative susceptibility mapping [14, 15] and water fraction estimation [17]. As the loss function in Eq. (8) is defined on generated data but not the physical evolution, this model can also be applied here. AM-IPADS enables the inverse learning from the image to physical parameters. But this learning does not mean that a sufficiently faithful results can be obtained. A possible way is to use the forward analytical model $\mathcal{F}$ to bridge the missed connection from physical parameters $\mathbf{p}$ to image $\mathbf{s} = \mathcal{F}(\mathbf{p})$, and then regularize the solution to has small errors of both physical parameters and images as:

$$\hat{\boldsymbol{\theta}} = \arg\min_{\boldsymbol{\theta}} \sum_{n=1}^{N} L(\mathbf{p}_n - \mathcal{N}(\mathbf{s}_n; \boldsymbol{\theta})) + L_{model}[\mathcal{F}(\mathbf{p}_n) - \mathcal{F}(\mathcal{N}(\mathbf{s}_n; \boldsymbol{\theta}))], \quad (10)$$

where $L_{model}$ is the additional loss with the analytical model. For example, in QSMnet[+] [15],

**p** is a susceptibility map, **s** is a local B$_0$ inhomogeneity image and $\mathcal{F}$ is an analytical differentiable dipole model defined in Eq. (4).

*C. Hybrid Approach*

Hybrid approach (HB-IPADS) means that IPADS learning integrates physical evolution and analytical model together. The former could be used for some parts that involve Bloch equation, specific MR pulse equations, and resonance structures. The latter is used to directly generate a large amount of training data from the analytical model with randomized parameters.

The HB-IPADS has been successfully applied to *in vivo* MRS spectra reconstruction [23, 24, 34], which considers the complex non-ideal acquisition conditions. The basis functions, $v_m(t)$ in Eq. (7), which are invariant to different subjects, are first generated using physical quantum mechanical simulations tools, such as jMRUI[1] and FID-A[2]. Then, all other physical signals, which are related to metabolite concentrations of objects, unexpected macromolecule signals and system imperfections, are directly generated according to the analytical expressions in Eq. (7), by varying parameters from an empirical range [23, 24, 34]. After minimizing the same loss function in Eq. (9), the trained network can be applied to MRS denoising [23, 34] and separation [24]. With an auto-encoder network, the mapping from the low signal-to-noise-ratio (SNR) spectra to the high SNR one is learnt on the IPADS data, and then applied to SNR improvement of realistic MRS [23]. Compared with the conventional spatial smoothness and subspace methods, auto-encoder reduces the mean square error of denoised $^{31}$P spectra on a numerical phantom by 90% and 58%, respectively [23].

Moreover, the HB-IPADS is also extended to MRI deep learning quantification, such as super paramagnetic iron oxide (SPIO) particle quantification [16]. This method firstly uses the analytical model in Eq. (4) to obtain the inhomogeneous local B$_0$ field of SPIO concentration, and then generates slice-modulated MRI image through Bloch simulation. The network learns the mapping from the generated images to the wavelet coefficients of spatial concentration distribution, which is finally obtained by performing the inverse wavelet transform. The network consists of encoder, bottleneck and four decoder sub-networks. This HB-IPADS method significantly outperforms traditional algorithms which are unreliable when the high concentration of SPIO has a large effect on the inhomogeneous field and phase.

---

[1] https://github.com/isi-nmr/jMRUI
[2] https://github.com/CIC-methods/FID-A

## IV. DATA ENHANCEMENT

Enhancing the IPADS data to fit for real applications is discussed here.

The experiment-specific characteristics, especially system imperfections, should be matched to real applications. For example, in CEST imaging, the inhomogeneity of $B_0$ has been carefully chosen for phantom (-0.4~0.4 ppm) and human skeletal muscle (-0.25~0.25 ppm), respectively [18]. Besides, a same range of imperfect parameters in training and testing data could greatly improve the deep learning performance. In real-time spiral MRI [35], a major limitation is image blurring introduced by off-resonance. A neural network trained with the same off-resonance range significantly increases the peak signal-to-noise ratio of deblurred images from 11.04 dB to 24.57 dB [35]. Furthermore, MRI images are usually acquired from multi-coils and the correlation of them can be reflected by the coil sensitivity maps. To adapt for real parallel imaging, these maps could be obtained on few real experimental data (5 subjects) from one scanner and then multiplied them onto single-channel images from other scanners or public databases [12].

The object-specific parameter should pay attention to the non-uniform distribution in real data. Many physical parameters of IPADS data are chosen from a given range with equal probability, resulting in the uniform distributions of these parameters in the training data [20]. This type of data generation provides unbiased representation when no other prior knowledge is provided. For example, uniform distributions of amplitudes and frequencies were made for IPADS spectrum training and have been successfully applied to many real protein spectra [20-22]. However, the reconstruction performance will be compromised if the true distribution is non-uniform. For example, many low-intensity peaks (the 2$^{nd}$ row of Fig. 4(c)) are lost in the reconstruction if they do not have a high ratio in the training data. Thus, a good estimation of the distribution for real data is valuable.

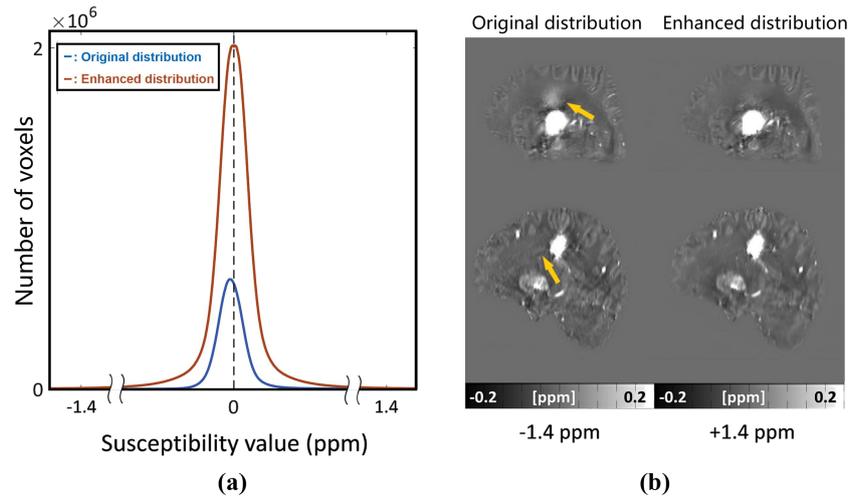

**Fig. 5**. Enhance the susceptibility distribution to a wider range [15]. (a) the original susceptibility distribution trained from healthy controls and the enhanced one. (b) Reconstructed QSM from hemorrhagic patients without and with data enhancement. The artifacts were marked by yellow arrow. ppm: parts per million.

In imaging, spatial and temporal distributions of physical parameters should be carefully tuned. In quantitative susceptibility mapping, the hemorrhagic lesions have higher susceptibility than healthy tissues. To better fit patient data, the susceptibility values were widened through multiplying different factors at different image regions of estimated maps, resulting in the much better artifacts removal [15] (Fig. 5). The random synthetic shapes assigned with restricted parametric values have been used for network training in OLED $T_2$ mapping [9-11] and water-fat imaging [25]. Instead, the spatial distribution with texture information from real data could be introduced into enhance IPADS learning. More realistic texture was achieved than the result of deep network trained with arbitrary random synthetic shapes [12, 25] (Fig. 6). In dynamic cardiac imaging, more motions were set at the beginning of the cycle than at the end, making the synthetic motion paths more realistic [26].

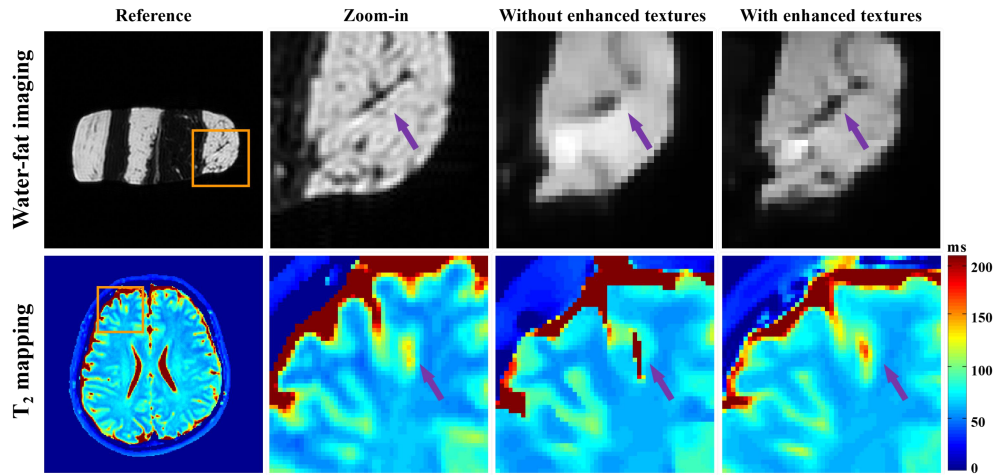

**Fig. 6**. Representative examples of spatial distribution enhanced IPADS learning. Upper row: Rich texture information from natural images were added into random shape-based templates for better water-fat imaging [19]. Lower row: More realistic textures were captured when real spatial distribution was introduced in $T_2$ mapping [8].

## V. ENHANCED LEARNING WITH IPADS

Narrowing the gap between IPADS and real data is a main task of the network design.

Regularizing the solution with general signal priors can generalize the IPADS network to practical data. These priors could be the smoothness [24] and low-rankness [21], that were widely adopted in traditional model-based reconstruction. A direct scheme is taking the output of a pre-trained network as an intermediate solution and then improving it with traditional optimization model. For example, the smoothness of image in MRSI is enforced with a finite-difference operator [24]. This scheme allows convergence characterization thus makes the algorithm more traceable. Another scheme is to design a network structure by imitating the iterative process in the model-based reconstruction, e.g., the low-rank MRS reconstruction (Fig. 4) [21]. This approach better preserves the low-intensity peaks and may handle the shifted distributions of spectra intensities in the training IPADS and target data. Besides, interpretable behavior of network, such as the progressive low-rank approximation, is also provided. Thus, the general signal prior is still powerful to improve IPADS network.

Refining the network parameters may repair knowledge missed in IPADS learning. Under the principle of meta-learning, introducing a sub-network to adjust hyper-parameters of the original network can adapt well to real data, enabling the network learning more knowledge beyond the IPADS training set [22]. For instance (Fig. 7), threshold in the backbone of sparse learning network can be adjusted so that each input had its own threshold to eliminate undersampling artifacts. Specifically, with the increase of the sampling rate of input signals (i.e., with the reduction of undersampling artifacts), the learned threshold decreases. Moreover, with more network phases, thresholds become smaller and artifacts are gradually removed. Thus, meta-learning can remedy the loss of reconstruction performance under unseen sampling rates in IPADS [22]. Another way is introducing few real data to adjust network parameters by transfer learning. This approach has been explored in conventional network training without IPADS [36]. For example, thousands of natural images were assembled with phase and coil information and then used to train MRI reconstruction network. By transfer learning from tens of real MRI images, nearly identical reconstruction performance is achieved as training thousands of real images [36]. Since IPADS learning has approximated physical laws in biomedical magnetic resonance, much less uncertainty needs to be addressed in the transfer learning. Thus, fewer real data may be required by IPADS learning than those of conventional network.

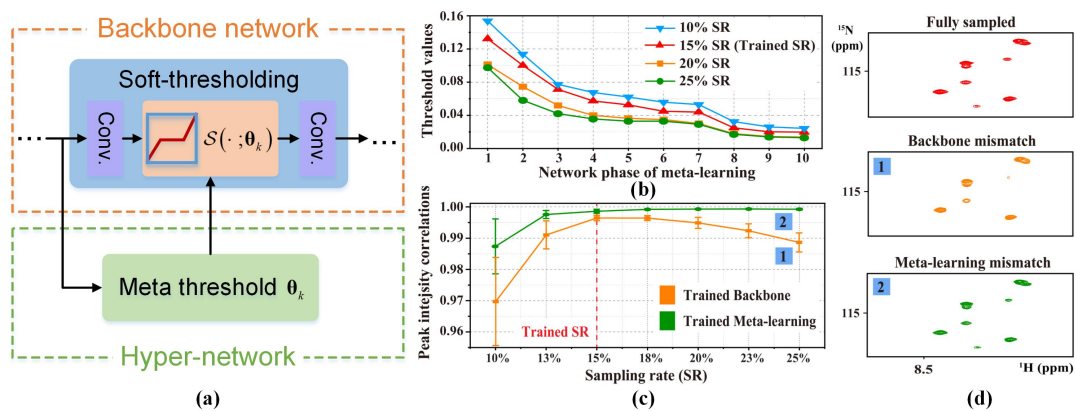

**Fig. 7**. Hyper-parameter adjustment with meta-learning in spectrum reconstruction [22]. (a) is the backbone network with soft-thresholding and the hyper-network that adjusts the hyper-parameter (threshold), (b) the threshold varies with different undersampling rates, (c) robust reconstruction to unseen sampling rates in IPADS, (d) a region of reconstructed spectra. Note: Higher peak intensity correlation means better reconstruction of spectral peaks.

Directly incorporating physical laws into network structure may improve the performance or relax the requirement of paired dataset. Due to the complex nature of MR signals, the operation in network, such as convolution, could be carefully treated and a complex form could reduce the reconstruction error [37]. Another good example is the recently proposed physics-informed neural networks for myocardial perfusion MRI quantification. By modeling the evolution of the concentration of contrast agent as differential equations, a physics-informed network is proposed to simultaneously estimate concentrations of contrast agent and quantify kinetic parameters [38]. This strategy does not need a large database of paired samples, but costs relatively long running time (1 hour per imaging slice) under a shallow network of 4 layers [38]. Since the solutions satisfy the underlying physical laws, the degrees of freedom to describe the imaging physics would be reduced. Thus, the requirement of paired data may be relaxed. Besides, the IPADS data could be used to train the initial network parameters offline. This would be very helpful to reduce the computational time and allows using deeper networks to improve parameter quantification performance.

## VI. OTHER MODALITIES LEARNING WITH IPADS

The application of IPADS has been applied successfully in other imaging modalities [5, 39-42]. For example, in doppler ultrasound imaging, the hemodynamic equations can be approximated with a physics-informed neural network [39] and the difference of predicted hemodynamic variables has been reduced by 20-50%. In endoscope imaging, a forward model was used to generate synthetic endoscope images [40], the synthetic-to-real gap was reduced by an adversarial generative network to remove patient-specific details and retain diagnostic information, resulting in an improved depth estimation performance by 78.7%. Thus, IPADS strategy is also applicable to other imaging modalities.

## VII. SUMMARY AND OUTLOOK

The deep learning has innovated the field of computational imaging. One of its bottlenecks is lacking of labels in challenging applications. This article reviews the recent progress of biomedical magnetic resonance on deep learning with imaging physics-based data synthesis (IPADS). Non or few real data is required since IPADS generate signals from partial differential equations or analytical solution models following physical laws, making the learning more scalable, explainable and better protecting privacy. Great potentials of IPADS have been evidenced in fast imaging, ultrafast signal reconstruction and accurate parameter quantification.

Mitigating the difference between synthetic with real data is still at the early stage. Many current methods highly depend on the accuracy of physical modeling and the selection of parameter ranges. Although good performance has been achieved on unseen real data, unpredictability in clinical situations remains the biggest risk of synthetic databased approaches. To generalize synthetic-data learning to real biomedicine and clinic, more effort should be made to mine the data/parameter distribution of real data, design robust networks and characterize the performance. Few subjective evaluations from radiologists have been found in the existed IPADS learning and needs to strengthen.

Besides, a powerful, open and any time accessible computing platform of IPADS is highly expected. As programming for physical signal evolution is non-trivial, an online platform that can fast and accurately generate data with user-defined parameters will allow more researchers to obtain huge IPADS datasets. These data can be used as training set for learning network or stress test data for developed deep learning imaging approaches. Scalable programming, such as graphic pulse sequence design for MRI, and high-precision simulation, e.g., non-rigid motion, proton diffusion and intravoxel multi-spin dephasing, will enable challenging imaging applications. It is worth noting that we have developed the prototype of such IPADS cloud platform (Fig. 8), called CloudBrain[1], and will continue to improve it with online physical evolution tools to better serve the community.

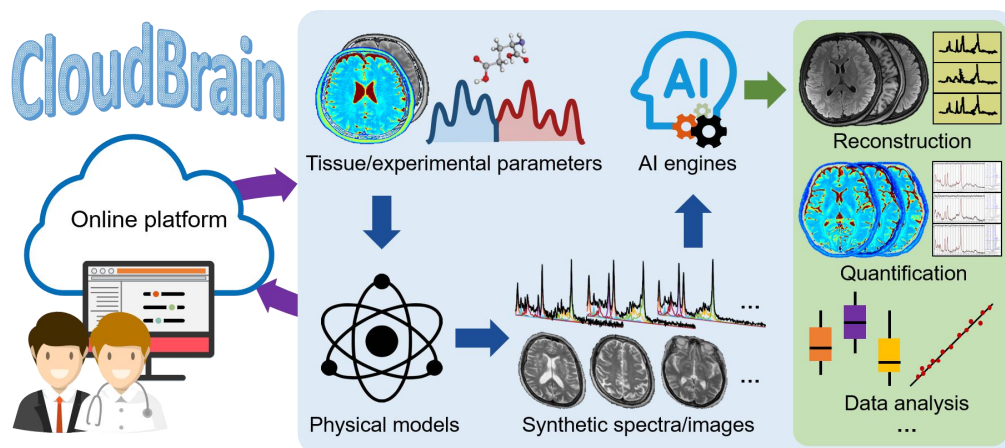

**Fig. 8**. A vision map of CloudBrain.

## ACKNOWLEDGMENTS

This work was supported in part by the National Natural Science Foundation of China (62122064, 61871341, 82071913), the Xiamen University Nanqiang Outstanding Talents Program. The authors are grateful to China Mobile for cloud computing resources. The authors thank Yirong Zhou, Jun Liu, Haoming Fang, Jiayu Li, Bangjun Chen for building a prototype of the CloudBrain; Haitao Huang, Xinran Chen, Chen Qian for providing experimental results and valuable discussions and guest editors and reviewers for providing

---

[1] https://csrc.xmu.edu.cn/XCloudAiImaging.html

valuable suggestions.

**REFERENCES**

[1] J. S. Duncan, M. F. Insana, and N. Ayache, "Biomedical imaging and analysis in the age of big data and deep learning," *Proc. IEEE,* vol. 108, no. 1, pp. 3-10, Jan. 2020. doi: 10.1109/jproc.2019.2956422.

[2] M. Jacob, J. C. Ye, L. Ying, and M. Doneva, "Computational MRI: Compressive sensing and beyond," *IEEE Signal Process. Mag.,* vol. 37, no. 1, pp. 21-23, Jan. 2020. doi: 10.1109/msp.2019.2953993.

[3] S. S. Wang, Z. H. Su, L. Ying, X. Peng, S. Zhu, F. Liang, D. G. Feng, and D. Liang, "Accelerating magnetic resonance imaging via deep learning," in *Proc. IEEE 13th Int. Symp. Biomed. Imag. (ISBI)*, pp. 514-517, Apr. 2016. doi: 10.1109/ISBI.2016.7493320.

[4] S. S. Wang, Z. W. Ke, H. T. Cheng, S. Jia, L. Ying, H. R. Zheng, and D. Liang, "DIMENSION: Dynamic MR imaging with both k-space and spatial prior knowledge obtained via multi-supervised network training," *NMR Biomed.,* vol. 35, no. 4, Sep. 2019. doi: 10.1002/nbm.4131.

[5] A. F. Frangi, S. A. Tsaftaris, and J. L. Prince, "Simulation and synthesis in medical imaging," *IEEE Trans. Med. Imag.,* vol. 37, no. 3, pp. 673-679, Mar. 2018. doi: 10.1109/tmi.2018.2800298.

[6] R. J. Chen, M. Y. Lu, T. Y. Chen, D. F. K. Williamson, and F. Mahmood, "Synthetic data in machine learning for medicine and healthcare," *Nat. Biomed. Eng.,* vol. 5, no. 6, pp. 493-497, Jun. 2021. doi: 10.1038/s41551-021-00751-8.

[7] O. Cohen, B. Zhu, and M. S. Rosen, "MR fingerprinting Deep RecOnstruction NEtwork (DRONE)," *Magn. Reson. Med.,* vol. 80, no. 3, pp. 885-894, Sep. 2018. doi: 10.1002/mrm.27198.

[8] O. Perlman, H. Ito, K. Herz, N. Shono, H. Nakashima, M. Zaiss, E. A. Chiocca, O. Cohen, M. S. Rosen, and C. T. Farrar, "Quantitative imaging of apoptosis following oncolytic virotherapy by magnetic resonance fingerprinting aided by deep learning," *Nat. Biomed. Eng.*, (Epub 2021 Nov 11), 2021. doi: 10.1038/s41551-021-00809-7.

[9] C. Cai, C. Wang, Y. Zeng, S. Cai, D. Liang, Y. Wu, Z. Chen, X. Ding, and J. Zhong, "Single-shot T2 mapping using overlapping-echo detachment planar imaging and a deep convolutional neural network," *Magn. Reson. Med.,* vol. 80, no. 5, pp. 2202-2214, Nov. 2018. doi: 10.1002/mrm.27205.

[10] J. Zhang, J. Wu, S. Chen, Z. Zhang, S. Cai, C. Cai, and Z. Chen, "Robust single-shot T2 mapping via multiple overlapping-echo acquisition and deep neural network," *IEEE Trans. Med. Imag.,* vol. 38, no. 8, pp. 1801-1811, Aug. 2019. doi: 10.1109/tmi.2019.2896085.

[11] S. M. Li, J. Wu, L. C. Ma, S. H. Cai, and C. B. Cai, "A simultaneous multi-slice T2 mapping framework based on overlapping-echo detachment planar imaging and deep learning reconstruction," *Magn. Reson. Med.,* vol. 87, no. 5, pp. 2239-2253, May. 2022. doi: 10.1002/mrm.29128.

[12] Q. Yang, J. Wang, J. Bao, X. Wang, L. Ma, Q. Yang, S. Cai, H. He, C. Cai, and J. Dong, "Model-based synthetic data-driven learning (MOST-DL): Application in single-shot T2 mapping with severe head motion using overlapping-echo acquisition," 2021. [Online]. Available: https://arxiv.org/abs/2107.14521.

[13] S. Gavazzi, C. A. T. van den Berg, M. H. F. Savenije, H. P. Kok, P. de Boer, L. J. A. Stalpers, J. J. W. Lagendijk, H. Crezee, and A. van Lier, "Deep learning-based reconstruction of in vivo pelvis conductivity with a 3D patch-based convolutional neural network trained on simulated MR data," *Magn. Reson. Med.,* vol. 84, no. 5, pp. 2772-2787, Nov. 2020. doi: 10.1002/mrm.28285.

[14] S. Bollmann, K. G. B. Rasmussen, M. Kristensen, R. G. Blendal, L. R. Ostergaard, M. Plocharski, K. O. Brien, C. Langkammer, A. Janke, and M. Barth, "DeepQSM - using deep learning to solve the dipole inversion for quantitative susceptibility mapping," *Neuroimage*, vol. 195, pp. 373-383, Jul. 2019. doi: 10.1016/j.neuroimage.2019.03.060.

[15] W. Jung, J. Yoon, J. Y. Choi, J. M. Kim, Y. Nam, E. Y. Kim, J. Lee, and S. Ji, "Exploring linearity of deep

neural network trained QSM: QSMnet+," *Neuroimage*, vol. 211, May. 2020. doi: 10.1016/j.neuroimage.2020.116619.

[16] G. Della Maggiora, C. Castillo-Passi, W. Qiu, S. Liu, C. Milovic, M. Sekino, C. Tejos, S. Uribe, and P. Irarrazaval, "DeepSPIO: Super paramagnetic iron oxide particle quantification using deep learning in magnetic resonance imaging," *IEEE Trans. Pattern Anal. Mach. Intell.*, vol. 44, no. 1, pp. 143-153, Jan. 2022. doi: 10.1109/tpami.2020.3012103.

[17] S. Jung, H. Lee, K. Ryu, J. E. Song, M. Park, W. J. Moon, and D. H. Kim, "Artificial neural network for multi-echo gradient echo-based myelin water fraction estimation," *Magn. Reson. Med.,* vol. 85, no. 1, pp. 394-403, Jan. 2021. doi: 10.1002/mrm.28407.

[18] L. Chen, M. Schar, K. W. Y. Chan, J. Huang, Z. Wei, H. Lu, Q. Qin, R. G. Weiss, P. C. M. van Zijl, and J. Xu, "In vivo imaging of phosphocreatine with artificial neural networks," *Nature Communications,* vol. 11, no. 1, Feb. 2020. doi: 10.1038/s41467-020-14874-0.

[19] C. L. Guo, J. Wu, J. T. Rosenberg, T. Roussel, S. H. Cai, and C. B. Cai, "Fast chemical exchange saturation transfer imaging based on PROPELLER acquisition and deep neural network reconstruction," *Magn. Reson. Med.,* vol. 84, no. 6, pp. 3192-3205, Dec. 2020. doi: 10.1002/mrm.28376.

[20] X. Qu, Y. Huang, H. Lu, T. Qiu, D. Guo, T. Agback, V. Orekhov, and Z. Chen, "Accelerated nuclear magnetic resonance spectroscopy with deep learning," *Angew. Chem. Int. Ed.,* vol. 59, no. 26, pp. 10297-10300, Jun. 2020. doi: 10.1002/anie.201908162.

[21] Y. Huang, J. Zhao, Z. Wang, V. Orekhov, D. Guo, and X. Qu, "Exponential signal reconstruction with deep hankel matrix factorization," *IEEE Trans. Neural Netw. Learn. Syst.*, (Epub 2021 Dec 23), 2021. doi: 10.1109/tnnls.2021.3134717.

[22] Z. Wang, D. Guo, Z. Tu, Y. Huang, Y. Zhou, J. Wang, L. Feng, D. Lin, Y. You, T. Agback, V. Orekhov, and X. Qu, "A sparse model-inspired deep thresholding network for exponential signal reconstruction--Application in fast biological spectroscopy," *IEEE Trans. Neural Netw. Learn. Syst.*, (Epub 2022 Feb 04), 2022. doi: 10.1109/tnnls.2022.3144580.

[23] F. Lam, Y. Li, and X. Peng, "Constrained magnetic resonance spectroscopic imaging by learning nonlinear low-dimensional models," *IEEE Trans. Med. Imag.,* vol. 39, no. 3, pp. 545-555, Mar. 2020. doi: 10.1109/tmi.2019.2930586.

[24] Y. Li, Z. Wang, R. Sun, and F. Lam, "Separation of metabolites and macromolecules for short-TE H-1-MRSI using learned component-specific representations," *IEEE Trans. Med. Imag.*, vol. 40, no. 4, pp. 1157-1167, Apr. 2021. doi: 10.1109/tmi.2020.3048933.

[25] X. Chen, W. Wang, J. Huang, J. Wu, L. Chen, C. Cai, S. Cai, and Z. Chen, "Ultrafast water-fat separation using deep learning-based single-shot MRI," *Magn. Reson. Med.*, (Epub 2022 Jan 31), 2022. doi: 10.1002/mrm.29172.

[26] M. Loecher, L. E. Perotti, and D. B. Ennis, "Using synthetic data generation to train a cardiac motion tag tracking neural network," *Med. Image Anal.,* vol. 74, Dec. 2021. doi: 10.1016/j.media.2021.102223.

[27] A. Loktyushin, K. Herz, N. Dang, F. Glang, A. Deshmane, S. Weinmuller, A. Doerfler, B. Scholkopf, K. Scheffler, and M. Zaiss, "MRzero - Automated discovery of MRI sequences using supervised learning," *Magn. Reson. Med.,* vol. 86, no. 2, pp. 709-724, Aug. 2021. doi: 10.1002/mrm.28727.

[28] Y. Zhang, K. Jiang, W. Jiang, N. Wang, A. J. Wright, A. Liu, and J. Wang, "Multi-task convolutional neural network-based design of radio frequency pulse and the accompanying gradients for magnetic resonance imaging," *NMR Biomed.,* vol. 34, no. 2, Feb. 2021. doi: 10.1002/nbm.4443.

[29] F. Bloch, W. W. Hansen, and M. Packard, "Nuclear induction," *Phys. Rev.,* vol. 69, no. 3-4, pp. 127-127, 1946.

[30] D. Ma, V. Gulani, N. Seiberlich, K. C. Liu, J. L. Sunshine, J. L. Duerk, and M. A. Griswold, "Magnetic

resonance fingerprinting," *Nature,* vol. 495, no. 7440, pp. 187-192, Mar. 2013. doi: 10.1038/nature11971.

[31] X. Qu, M. Mayzel, J. F. Cai, Z. Chen, and V. Orekhov, "Accelerated NMR spectroscopy with low-rank reconstruction," *Angew. Chem. Int. Ed.,* vol. 54, no. 3, pp. 852-854, Jan. 2015. doi: 10.1002/anie.201409291.

[32] C. B. Cai, M. J. Lin, Z. Chen, X. Chen, S. H. Cai, and J. H. Zhong, "SPROM - an efficient program for NMR/MRI simulations of inter- and intra-molecular multiple quantum coherences," *C. R. Phys.,* vol. 9, no. 1, pp. 119-126, Jan. 2008. doi: 10.1016/j.crhy.2007.11.007.

[33] F. Liu, J. V. Velikina, W. F. Block, R. Kijowski, and A. A. Samsonov, "Fast realistic MRI simulations based on generalized multi-pool exchange tissue model," *IEEE Trans. Med. Imag.,* vol. 36, no. 2, pp. 527-537, Feb. 2017. doi: 10.1109/tmi.2016.2620961.

[34] H. H. Lee, and H. Kim, "Intact metabolite spectrum mining by deep learning in proton magnetic resonance spectroscopy of the brain," *Magn. Reson. Med.,* vol. 82, no. 1, pp. 33-48, Jul. 2019. doi: 10.1002/mrm.27727.

[35] Y. Lim, Y. Bliesener, S. Narayanan, and K. S. Nayak, "Deblurring for spiral real-time MRI using convolutional neural networks," *Magn. Reson. Med.*, vol. 84, no. 6, pp. 3438-3452, 2020. doi: 10.1002/mrm.28393.

[36] S. U. Dar, M. Ozbey, A. B. Catli, and T. Cukur, "A transfer-learning approach for accelerated MRI using deep neural networks," *Magn. Reson. Med.,* vol. 84, no. 2, pp. 663-685, Aug. 2020. doi: 10.1002/mrm.28148.

[37] S. S. Wang, H. T. Cheng, L. Ying, T. H. Xiao, Z. W. Ke, H. R. Zheng, and D. Liang, "DeepcomplexMRI: Exploiting deep residual network for fast parallel MR imaging with complex convolution," *Magn. Reson. Imaging*, vol. 68, pp. 136-147, May. 2020. doi: 10.1016/j.mri.2020.02.002.

[38] R. L. M. van Herten, A. Chiribiri, M. Breeuwer, M. Veta, and C. M. Scannell, "Physics-informed neural networks for myocardial perfusion MRI quantification," *Med. Image Anal.,* vol. 78, pp. 102399, May. 2022. doi: 10.1016/j.media.2022.102399.

[39] M. Sarabian, H. Babaee, and K. Laksari, "Physics-informed neural networks for brain hemodynamic predictions using medical imaging," *IEEE Trans. Med. Imag.*, (Epub 2022 Mar 23), 2022. doi: 10.1109/TMI.2022.3161653.

[40] F. Mahmood, R. Chen, and N. J. Durr, "Unsupervised reverse domain adaptation for synthetic medical images via adversarial training," *IEEE Trans. Med. Imag.*, vol. 37, pp. 2572-2581, Dec. 2018. doi: 10.1109/TMI.2018.2842767.

[41] F. Mahmood, R. Chen, S. Sudarsky, D. Yu, and N. J. Durr, "Deep learning with cinematic rendering: fine-tuning deep neural networks using photorealistic medical images," *Phys. Med. Biol.*, vol. 63, no. 18, pp. 185012, Sep. 2018. doi: 10.1088/1361-6560/aada93.

[42] K. Incetan, I. O. Celik, A. Obeid, G. I. Gokceler, K. B. Ozyoruk, Y. Almalioglu, R. J. Chen, F. Mahmood, H. Gilbert, N. J. Durr, and M. Turan, "VR-Caps: A virtual environment for capsule endoscopy," *Med. Image Anal.,* vol. 70, pp. 101990, May. 2021. doi: 10.1016/j.media.2021.101990.